\documentclass[conference]{IEEEtran}
\IEEEoverridecommandlockouts
\usepackage{cite}
\usepackage{amsmath,amssymb,amsfonts}
\usepackage{algorithmic}
\usepackage{graphicx}
\usepackage{textcomp}
\usepackage{xcolor}
\def\BibTeX{{\rm B\kern-.05em{\sc i\kern-.025em b}\kern-.08em
    T\kern-.1667em\lower.7ex\hbox{E}\kern-.125emX}}
\begin{document}

\title{A Light-weight Deep Learning Model for Remote Sensing Image Classification
}

\author{\IEEEauthorblockN{Lam Pham*}
\IEEEauthorblockA{\textit{Austrian Institute of Technology} \\
Vienna, Austria \\
lam.pham@ait.ac.at}
\and
\IEEEauthorblockN{Cam Le*}
\IEEEauthorblockA{\textit{HCM University of Technology} \\
HCM, VietNam \\
cam.levt123@hcmut.edu.vn }
\and
\IEEEauthorblockN{Dat Ngo}
\IEEEauthorblockA{\textit{University of Essex} \\
Colchester, UK \\
dn22678@essex.ac.uk} \\
\and
\IEEEauthorblockN{Anh Nguyen}
\IEEEauthorblockA{\textit{FPT Soft Company} \\
HCM, VietNam \\
AnhNTN34@fsoft.com.vn}
\and
\IEEEauthorblockN{Jasmin Lampert}
\IEEEauthorblockA{\textit{Austrian Institute of Technology} \\
Vienna, Austria \\
Jasmin.Lampert@ait.ac.at}
\and
\IEEEauthorblockN{Alexander Schindler}
\IEEEauthorblockA{\textit{Austrian Institute of Technology} \\
Vienna, Austria \\
Alexander.Schindler@ait.ac.at}
\and
\IEEEauthorblockN{Ian McLoughlin}
\IEEEauthorblockA{\textit{Singapore Institute of Technology} \\
Singapore \\
ian.mcloughlin@singaporetech.edu.sg}


\thanks{(*) Lam Pham and Cam Le made equal contribution to this paper.}%

}

\maketitle

\begin{abstract}
In this paper, we present a high-performance and light-weight deep learning model for Remote Sensing Image Classification (RSIC), the task of identifying the aerial scene of a remote sensing image.
To this end, we first evaluate various benchmark convolutional neural network (CNN) architectures: MobileNet V1/V2, ResNet 50/151V2, InceptionV3/InceptionResNetV2, EfficientNet B0/B7, DenseNet 121/201, ConNeXt Tiny/Large.
Then, the best performing models are selected to train a compact model in a teacher-student arrangement. 
The knowledge distillation from the teacher aims to achieve high performance with significantly reduced complexity.
By conducting extensive experiments on the NWPU-RESISC45 benchmark, our proposed teacher-student models outperforms the state-of-the-art systems, and has potential to be applied on a wide rage of edge devices.  
\end{abstract}


\section{Introduction}
Remote sensing image classification (RSIC) is a core task for a range of real-world applications including land use classification, natural hazard assessment~\cite{poursanidis2017remote}, scene-driven geospatial object detection~\cite{feng2015uav}, and environmental monitoring~\cite{van2013remote}. The task has therefore drawn much attention from the research community in recent years, including in the area of datasets and benchmarks.
The earliest RSIC dataset, UCM~\cite{yang2010bag}, was proposed in 2010. 
Subsequently, more challenging RSIC datasets have been published, such as NWPU VHR-10 (2014)~\cite{cheng2014multi}, SAT6 (2015)~\cite{basu2015deepsat}, SIRI-WHU (2015)~\cite{zhao2015dirichlet}, AID (2017)~\cite{xia2017aid}, OPTIMAL (2018)~\cite{wang2018scene}, NWPU-RESISC45 (2017)~\cite{cheng2017remote}, etc. 
Among these published datasets, NWPU-RESISC45 has the largest number of classes, comprising 45 image scenes, each of which is represented by 700 remote sensing images.  
Additionally, a wide range of classification models have been published for RSIC tasks.
Early systems used conventional image processing techniques such as Texture Descriptors (TD)~\cite{manjunath2001color}, Local binary patterns (LBP)~\cite{pietikainen2010local}, Color Histogram (CH), Histogram of Oriented Gradient (HOG)~\cite{dalal2005histograms}, Scale-Invariant Feature Transformation (SIFT)~\cite{lowe2004sift} to extract hand-crafted features. 
Then, these features were classified by traditional machine learning based models such as Support Vector Machine (SVM)~\cite{du2012multiple, cheng2017remote}, Gaussian Mixture Model (GMM)~\cite{gmm_model}, etc. 
More recently proposed RSIC systems leveraged deep learning based network architectures, which have proven to be more effective compared to traditional machine learning methods~\cite{mehmood2022remote, gu2019survey}.
Most deep learning based systems for RSIC make use of Convolutional Neural Network (CNN) based architectures such as ResNet~\cite{shabbir2021satellite}, DenseNet~\cite{tong2020channel}, EfficientNet~\cite{zhang2020transfer} or Transformer~\cite{sota_13}. 
Although deep learning based RSIC systems have demonstrated the potential for very good  performance~\cite{wang2022empirical}, these network architectures involve large footprint models with a high number of trainable parameters~\cite{wang2022empirical}.
This causes challenges to apply such deep learning based RSIC models within edge devices~\cite{sun2021mind}. 
In this paper, we aim to develop a low footprint RSIC model which is capable of achieving high-performance by leveraging the strength of advanced high complexity models to achieve cutting-edge performance. The resulting distilled student architecture achieves a model size reduction of 98\% at the cost of a 1.4\% relative drop in performance.
Our main contributions are as follows:\\
(a) A mechanism to combine individual high-performing CNN-based networks trained on the RSIC task, to inform a single robust teacher network. Given the teacher, we apply a teacher-student scheme to train the student.
Using knowledge distillation from the teacher, the student not only performs well but is also a low complexity model.
In this paper, we propose a constraint of maximum 5 million trainable parameters for a low-complexity RSIC model. 
This is consistent with the capability of typical edge devices.
(b) We evaluate our proposed teacher and student models on the NWPU-RESISC45 benchmark~\cite{cheng2017remote}. 
Results reveal that the proposed models outperform state-of-the-art systems with or without considering the issue of complexity -- demonstrating the ability of the technique to enable implementation on a range of edge devices. 

\section{The three-phase process to develop and achieve a high-performance and low-complexity RSIC model}
In this section, we describe the methods employed to achieve a high-performance and low-complexity RSIC model, which leverages a teacher-student arrangement~\cite{gou2021knowledge}.
In particular, the process comprises three main phases:
\begin{itemize}
    \item Phase I: We first evaluate a wide range of benchmark convolution neural network (CNN) based architectures. Then, we select which networks (i.e. the best performance models) to use for developing the teacher model, and which network is used for the student model (i.e. the student model not only performs well but also presents a low footprint).
    \item Phase II: In this phase, the best performance models from Phase I are used to develop the teacher. After training the proposed teacher, the feature maps at the next to last dense layer of the teacher are extracted. The extracted feature maps are referred to as high-level features.
    \item Phase III: Finally, the student network, which selected in Phase I, is trained with the high-level features (i.e. via knowledge distillation from the teacher) to achieve a high-performance and low-complexity RSIC model.
\end{itemize}

\subsection{Phase I: Evaluate the benchmark neural networks to select high-performance networks for the teacher and student}\label{phase-1}

\begin{figure}[t]
    \centering
    \includegraphics[width=1.0\linewidth]{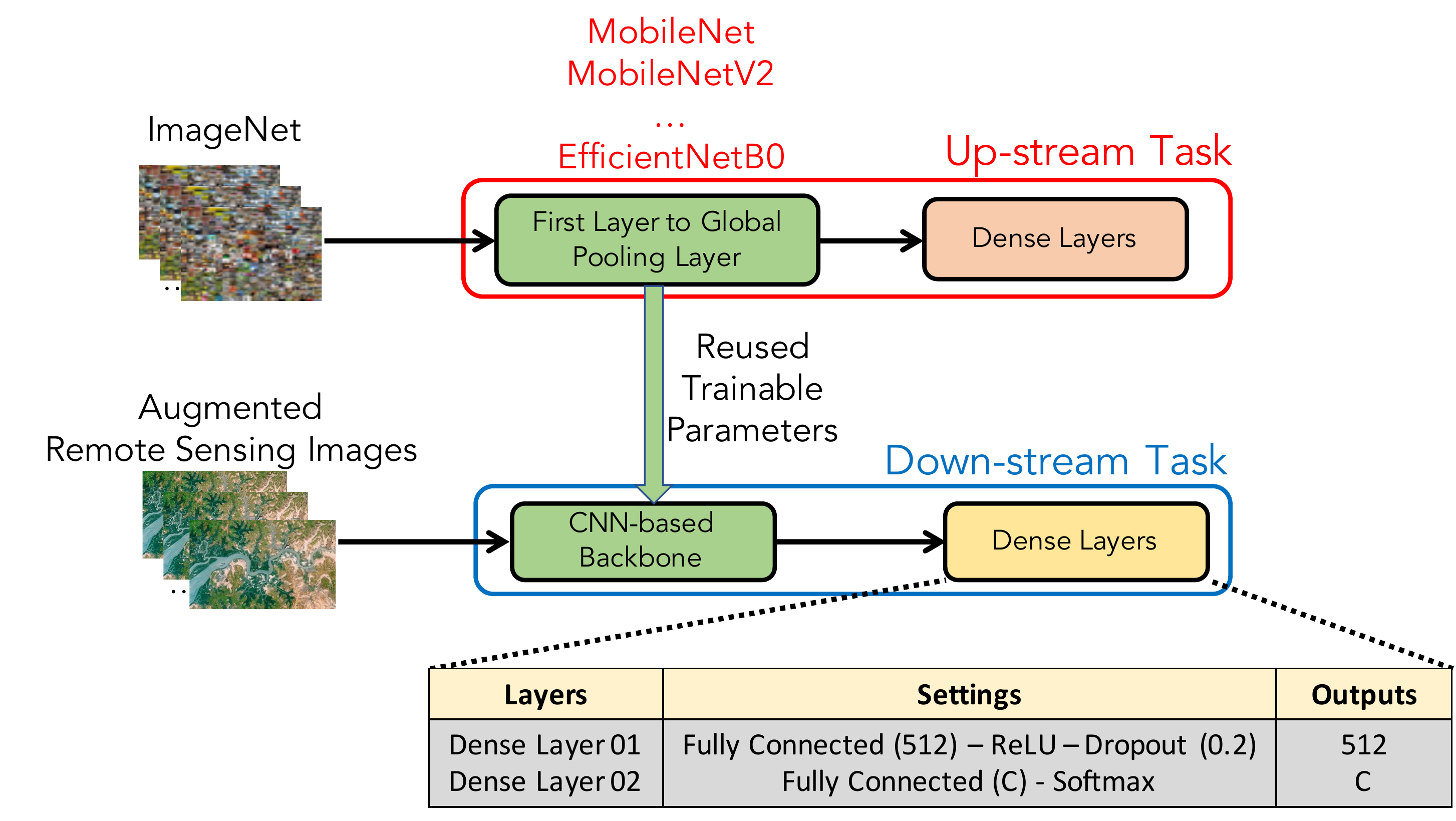}
       	\vspace{-0.7cm}
	\caption{Evaluation of benchmark networks using the transfer learning technique.}
    \label{fig:phase_1}
\end{figure}

We assessed various convolutional neural network (CNN) based architectures for both the teacher and the student models by evaluating twelve different benchmark deep convolutional neural networks: MobileNet, MobileNetV2, ResNet50, Resnet151V2, InceptionV3, InceptionResNetV2, DenseNet121, DenseNet201, EfficientNetB0, EfficientNetB7, ConvNeXtTiny, and ConvNeXtLarge, all available in the Keras library~\cite{keras_app}.
As the top of Figure~\ref{fig:phase_1} shows, the benchmark networks are first trained with the ImageNet dataset~\cite{Imagenet}, referred to as the up-stream task.
Then, the first layer to the global pooling layer of these pre-trained networks are extracted and combined with a Dense Layers block to perform the down-stream RSIC task as shown at the bottom of Figure~\ref{fig:phase_1}.
In other words, we apply a transfer learning method in which the first layer to the global pooling layer, trained from the up-stream task using the ImageNet dataset~\cite{Imagenet}, are transferred to the down-stream RSIC task.
The Dense Layers block is considered to house the adapting layers for the down-stream RSIC task.

We also apply data augmentation for the RSIC down-stream task, namely Image Rotation~\cite{rotation_aug} and Mixup~\cite{mixup1}, performed on the remote sensing image input dataset.
In particular, all images in an original RSIC dataset are rotated using three different angles: 90, 180, and 270\textdegree. 
Since three angles are used, the augmented dataset is four times larger than the original.
Next, batches of 60 images are randomly selected from the new dataset.  
For each batch, we apply the Mixup~\cite{mixup1} method to mix the images within one batch with random ratios.
Both Uniform and Beta distributions are used to generate the mixup ratios, and we make use of both the rotation augmented image database in addition to the new mixup images; as a consequence the batch size increases by three times from 60 to 180 images.

Thanks to the use of Mixup~\cite{mixup1} for data augmentation, the labels will no longer be in one-hot encoding format. 
Therefore we apply Kullback-Leibler divergence (KL) loss~\cite{kl_loss} instead of Entropy loss to train the evaluating models, as in equation~\ref{eq:kl_loss}:
\begin{align}
   \label{eq:kl_loss}
   Loss_{KL}(\Theta) = \sum_{b=1}^{B}\sum_{c=1}^{C}\mathbf{y}_{bc}\log \left\{ \frac{\mathbf{y}_{bc}}{\mathbf{\hat{y}}_{bc}} \right\}  +  \frac{\lambda}{2}||\Theta||_{2}^{2} \,,
\end{align}
where \(\Theta\) presents trainable parameters, the constant \(\lambda\) is empirically set to $0.0001$, the batch size $B$, the number of classes C, $\mathbf{y_{bc}}$ and $\mathbf{\hat{y}_{bc}}$ denote expected and predicted probabilities of an input image, respectively.
Note that we set the low learning rate to be 0.0001 and none of trainable parameters are frozen during the training process.

\subsection{Phase II: Develop the teacher and extract high-level features from the teacher}\label{phase-2}
\begin{figure}[t]
    \centering
    \includegraphics[width=1.0\linewidth]{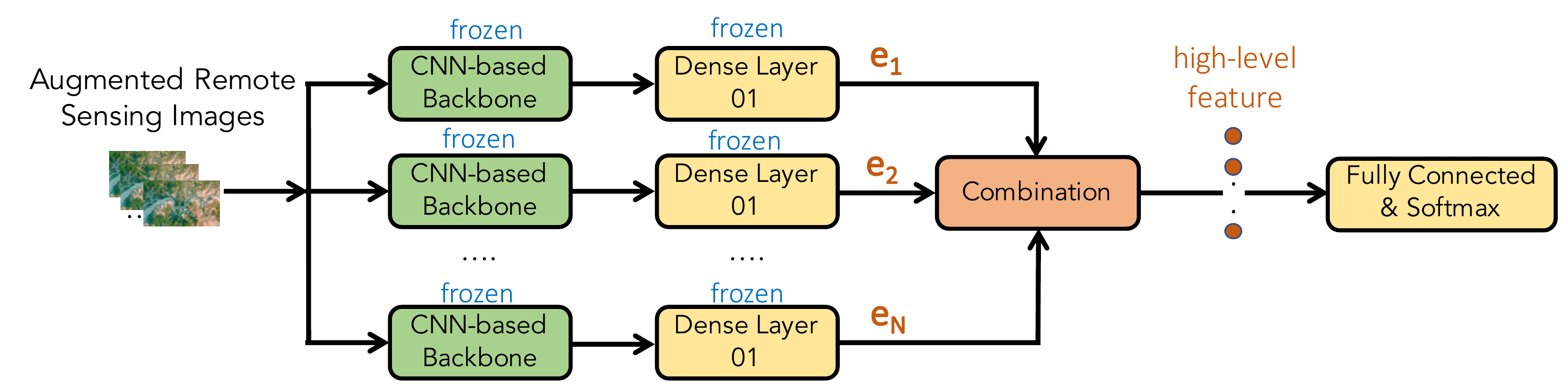}
           	\vspace{-0.7cm}
	\caption{The Teacher model generated from individual high-performance networks}
    \label{fig:phase_2}
\end{figure}

Given $N$ high-performance models selected from Phase I, we then develop and train the teacher architecture during this phase. 
Again, we leverage parameter based transfer learning techniques to develop the teacher as shown in Figure~\ref{fig:phase_2}.
In particular, the first layer to the Dense Layer 01 of Dense Layers block from all $N$ high-performance networks described in Phase I are reused and then combined to generate a composite high-level feature.
If we consider $N$ vectors $\mathbf{e_{n}}\in R^{512}$ as the output of the Dense Layer 01, the Combination block used to generate the composite high-level feature in Figure~\ref{fig:phase_2} by, 
\begin{align}
   \label{eq:custom-layer}
   f(\mathbf{e_{1}, e_{2}, ...,e_{N}}) = \mathbf{\sum_{n=1}^{N} e_{n}}\odot\mathbf{w_{i}} + \mathbf{b}
\end{align}

where $\mathbf{w_{i}}, \mathbf{b}\in R^{512}$ are weight and bias trainable parameters.  
The high-level feature is finally transferred into a Fully Connected layer followed by a Softmax for classifying to target classes.
When we finish training the teacher model, the high-level features are then extracted and used for the knowledge distillation process to train the student in Phase III which follows.

Data augmentation is used when training the teacher network, however only Image Rotation~\cite{rotation_aug} is applied at this and thus the labels can remain in one-hot format, and Entropy loss can be used to train the teacher model as in equation~\ref{eq:en_loss}: 
\begin{align}
   \label{eq:en_loss}
   Loss_{E}(\Theta) = -\sum_{b=1}^{B}\sum_{c=1}^{C}\mathbf{y}_{bc}\log \left\{ \mathbf{\hat{y}}_{bc} \right\}  +  \frac{\lambda}{2}||\Theta||_{2}^{2} \,,
\end{align}
where  \(\Theta\) are trainable parameters, the constant \(\lambda\) is set to $0.0001$, the batch size $B$ and the number of classes C, $\mathbf{y_{bc}}$ and $\mathbf{\hat{y}_{bc}}$ denote expected and predicted probabilities of a particular image, respectively.

We again set the low learning rate to 0.0001, and the trainable parameters of the first layer to the Dense Layer 01 are frozen when training the teacher.
In other words, only trainable parameters in the Combination block and in the finally Fully Connected layer are updated during the training process.

\subsection{Phase III: Train the student network to achieve high-performance and low-complexity RSIC}\label{phase-3}
\begin{figure}[t]
    \centering
    \includegraphics[width=1.0\linewidth]{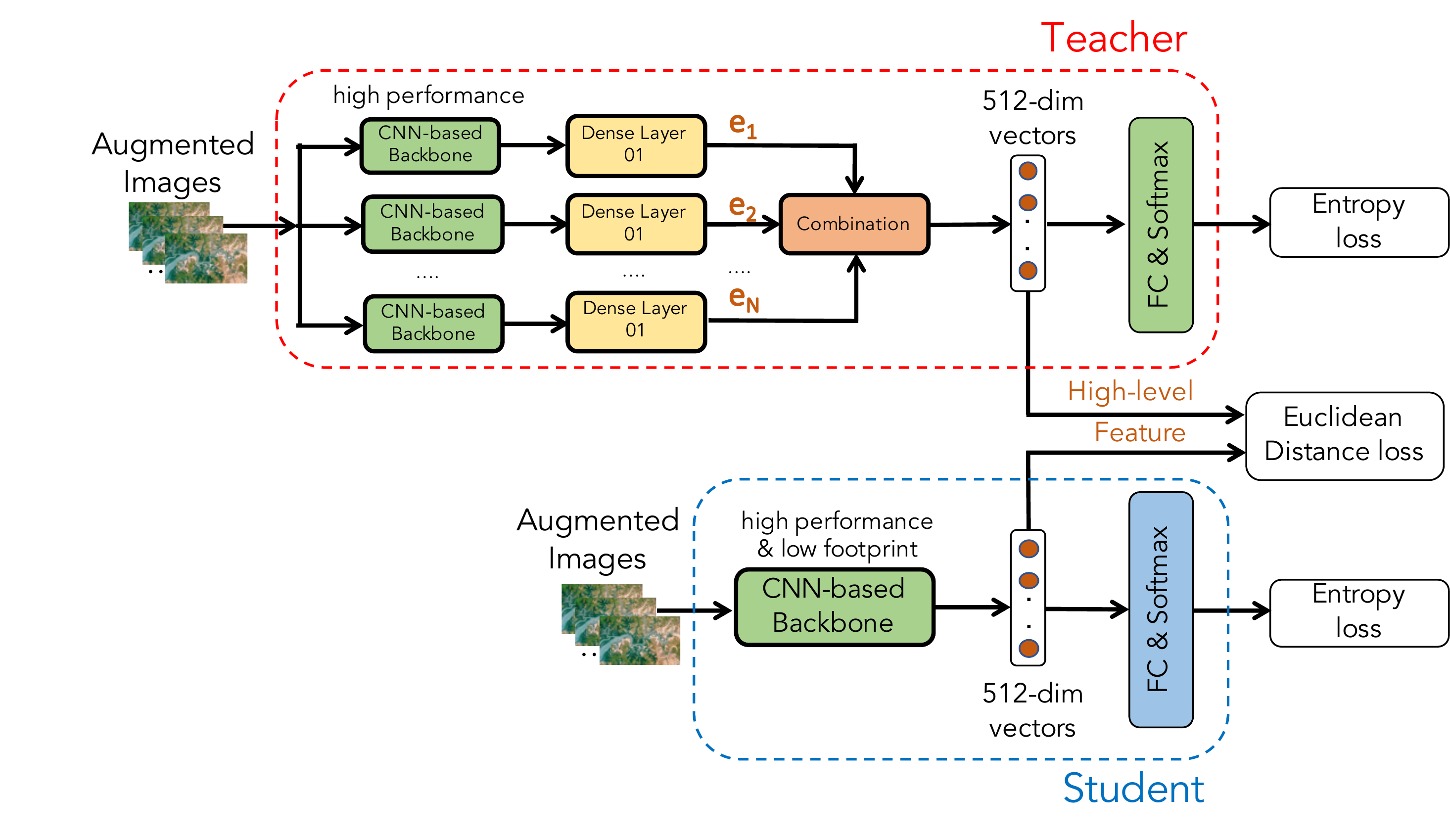}
           	\vspace{-0.9cm}
	\caption{The student model with knowledge distillation from the teacher.}
    \label{fig:phase_3}
\end{figure}

From the results in Phase I, a network architecture, which not only performs well but also presents a low footprint, is selected and considered as the base student model.
We then train the student with the high-level features extracted from the teacher mentioned in Phase II.
As Figure~\ref{fig:phase_3} shows, the student is trained with two loss functions.
While the first Entropy loss is used for the classification task, the Euclidean Distance loss helps to ensure the high-level features of the student become closer to the high-level features extracted from the teacher, effectively guiding the feature discrimination ability of the student.
The ratio between both losses is empirically set to 0.5/0.5.

Regarding the data augmentation used to train the student, only Image Rotation~\cite{rotation_aug} is applied.
We also set the low learning rate of 0.0001 and no trainable parameters are frozen during the student training phase.
 
\section{Experiments and results}
\begin{table*}[t]
    \caption{Performance comparison of benchmark CNN based network architectures on \\ the NWPU-RESISC45 task with a training/testing split of 20/80.} 
        	\vspace{-0.2cm}
    \centering
    \scalebox{1.0}{

    \begin{tabular}{|c| c|c| c|c| c|c|}  
        \hline 
        \textbf{Network}   &\textbf{MobileNetV2} &\textbf{MobileNet}    &\textbf{ResNet50}       &\textbf{Resnet151V2}      &\textbf{InceptionV3}  &\textbf{InceptionResNetV2} \\
        \hline 
        \textbf{Accuracy (\%)}      &88.0  &90.8  &91.8  &92.4  &86.9  &90.5       \\
        \textbf{Parameters (M)}     &2.9   &3.7   &24.6  &59.2  &22.8  &55.1       \\  
        \hline 
        \hline 

        \textbf{Network}  &\textbf{DenseNet121}  &\textbf{DenseNet201}  &\textbf{EfficientNetB0} &\textbf{EfficientNetB7}   &\textbf{ConvNeXtTiny}  &\textbf{ConvNeXtLarge}\\
        \hline 
        \textbf{Accuracy (\%)}      &92.0  &\textbf{93.3}  &92.3  &\textbf{93.6}  &93.0  &\textbf{95.3}     \\
        \textbf{Parameters (M)}     &7.5   &19.1  &4.7   &65.1  &27.5  &196.6    \\  
        \hline 

      \end{tabular}
    }
    \label{table:res_01} 
\end{table*}

\begin{table}[t]
	\caption{Performance comparison of the Teacher (a combination of ConvNeXtLarge,DenseNet201,EfficientNetB7), the student (EfficientNetB0) with various settings, on the NWPU-RESISC45 task  with a training/testing split of 20/80.} 
        	\vspace{-0.2cm}
    \centering
    \scalebox{1.0}{

    \begin{tabular}{|l| c|c| c|}  
        \hline 
        \textbf{Network}   &\textbf{Accuracy (\%)} &\textbf{Parameters (M)}    \\
        \hline 
        \textbf{Teacher}                       &96.2  &280.8  \\
        \textbf{EfficientNetB0 (student)}                &92.3  &4.7    \\
        \textbf{EfficientNetB0+distillation}        &94.8  &4.7    \\
        \textbf{EfficientNetB0-6B+distillation}     &94.4  &3.0    \\  
        \textbf{EfficientNetB0-5B+distillation}     &93.5  &0.93   \\  
        \textbf{EfficientNetB0-4B+distillation}     &91.3  &0.37   \\  
        \textbf{EfficientNetB0-3B+distillation}     &85.6  &0.11   \\  
        \hline 

      \end{tabular}
    }
    \label{table:res_02} 
\end{table}
\begin{table}[t]
    \caption{Performance (Acc.\%) comparison of the propose teacher against state-of-the-art RSIC systems on the NWPU-RESISC45 benchmark with two split arrangements, and without any trainable parameter size constraint.} 
  	\vspace{-0.2cm}
    \centering
    \scalebox{0.9}{
    \begin{tabular}{|l|c|c|} 
        \hline 
\textbf{Methods}                                       &\textbf{10\% training} &\textbf{20\% training} \\
	    \hline         
MG-CAP  \cite{sota_01}                &90.8         &93.0 \\ 
EfficientNet-B3-aux \cite{sota_05}    &91.1                   &93.8  \\

ResNeXt-101 + MTL \cite{sota_11}   &91.9                   &94.2 \\
MBLANet  \cite{chen2021remote}                     &92.3                   &94.6 \\ 
GRMANet \cite{li2021gated}                          &93.2                   &94.7 \\ 
KFBNet \cite{li2020high}        & 93.1    & 95.1 \\ 
CTNet \cite{deng2021cnns}      & 93.9  & 95.4 \\ 
TRS \cite{sota_13}                    &93.1                   &95.6 \\ 
RSP-ViTAEv2-S-E100 \cite{wang2022empirical}                    &94.4                   &95.6 \\ 

        \hline 
\textbf{Our system (Teacher)}                           &\textbf{94.6}          &\textbf{96.2} \\
         \hline 
      \end{tabular}
    }
    \label{table:res_03} 
\end{table}

\begin{table}[t]
    \caption{Performance comparison of the proposed student against state-of-the-art systems on the benchmark NWPU-RESISC45 dataset with two split settings and a constraint of no more than 5M trainable parameters.} 
   	\vspace{-0.2cm}
    \centering
    \scalebox{0.9}{
    \begin{tabular}{|l|c|c|} 
        \hline 
\textbf{Methods}                                       &\textbf{10\% training} &\textbf{20\% training} \\
	    \hline         
EfficientNet-B0-aux ($\approx$ 5M) \cite{sota_05}   &90.0                   &92.9  \\
DMP-MobileNetV2 (3.47 M) \cite{hou2022dmpconv}                     &90.3                   &93.1 \\
BiMobileNet (2.52 M) \cite{yu2020efficient}                     &91.9                   &93.9 \\
SE-MDPMNet (5.17 M) \cite{sota_10}                     &91.8                   &94.1 \\
LGRIN (4.63 M) \cite{sota_12}                          &91.9                   &94.4 \\

        \hline 
\textbf{Our system (Student+distillation)}                           &\textbf{93.3}          &\textbf{94.8} \\
         \hline 
      \end{tabular}
    }
    \label{table:res_04} 
\end{table}

\subsection{Dataset}
\label{dataset}
In this paper, the benchmark dataset of NWPU-RESISC45\cite{cheng2017remote} is used to evaluate all state of the art and proposed models.
The dataset, which was collected from more than 100 different countries and regions around the world, consists of 31,500 remote sensing images separated into 45 scene classes. 
Each class comprises 700 RGB images with a resolution of $256\times256\times3$.
To compare with state-of-the-art systems, we comply with the original settings mentioned in~\cite{cheng2017remote}.
We then split the NWPU-RESISC45 dataset into Training and Testing sets with two different ratios: 10\%-90\% and 20\%-80\%, respectively.

\subsection{Evaluation metric}
\label{metric}
As the Accuracy (Acc.\%) has been used as the main metric to compare performance among the RSIC systems, we also apply the metric in this paper.
Additionally, as we aim to achieve a low complexity model for the RSIC task, we compute the number of trainable parameters (M) to compare against state-of-the-art RSIC systems. 

\subsection{Experimental settings}
\label{setting}
We constructed our proposed deep learning networks with Tensorflow using the Adam method~\cite{adam} for optimization.
The training and evaluating processes are conducted on two Titan RTX 24GB GPUs.
The results presented in this paper are all the average scores from 10 individual experimental runs.

\subsection{Results and Discussions}

As experimental results show in Table~\ref{table:res_01}, we can see that ConvNeXt, EfficientNet, DenseNet based models are competitive and outperform MobileNet, ResNet and Inception based models. 
Particularly, the best network architectures of ConvNeXtLarge and ConvNeXtTiny achieve 95.3\% and 93.0\% accuracy, respectively.
Around 2\% worse than ConvNeXtLarge, the performance of EfficientNetB7 and DenseNet201 on the NWPU-RESSIC45 task are 93.6\% and 93.3\%, respectively.
Meanwhile, their smaller variants named DenseNet121 and EfficientNetB0 achieve over 92\% accuracy. 

Although ConvNeXt, EfficientNet and DenseNet based models perform well among the evaluating network architectures, these involve large footprints.
In particular, the three best variants, namely ConvNeXtLarge, EfficientNetB7, and DensNet201 have some of the largest parameter set sizes of 196.6, 65.1, and 19.1 M, respectively.
Among the ConvNeXt, EfficientNet and DenseNet variants, only EfficientNetB0 combines a relatively good accuracy of 92.3\% with a low complexity footprint (4.7 M parameters).
As a result, we select EfficientNetB0 as the foundation network for the student model required in Phase III.
We also note that DenseNet201, EfficientNetB7 and ConvNeXtLarge perform better than 93\% and their general architectures are dissimilar to each other.
We therefore, select these three network architectures to generate the teacher, as required in Phase II.

As Table~\ref{table:res_02} shows, the teacher (i.e. a combination of DenseNet201, EfficientNetB7 and ConvNeXtLarge) achieves an accuracy of 96.2\%, but with a very large footprint of 280.8 M parameters. 
Knowledge distillation from this capable teacher into student EfficientNetB0 allows it to achieve an accuracy of 94.8\% while maintaining a low complexity of 4.7 M parameters.
To propose a wide range of low complexity models, we further evaluate variants of the student EfficientNetB0 model.
In particular, variants of the student are generated by removing certain convolutional blocks in the EfficientNetB0 backbone architecture to reduce complexity further.
EfficientNetB0-6B to EfficientNetB0-3B are variants of EfficientNetB0 obtained by removing convolutional block 7 only, removing convolutional blocks 6 and 7, removing all convolutional blocks from 5 to 7 and removing all convolutional blocks 4 to 7 inclusive. 
Experimental results in Table~\ref{table:res_02} indicate that when the footprint of EfficientB0 based students is reduced, the accuracy performance also tends to decrease.
However, we can achieve a very low complexity model of 0.37 M parameters with a performance of 91.3\% from EfficientNetB0-4B, which opens the potential for RSIC applications on a very wide range of edge devices.

Finally, we compare our proposed models to the state-of-the-art RSIC systems basing on two criteria: (1) accuracy performance without any model complexity constraint and (2) accuracy performance with a constraint of 5 M trainable parameters maximum. 
As Table~\ref{table:res_03} shows, RSIC performance with the first criterion reveals that our proposed teacher (i.e. a combination of ConvNeXtLarge, DenseNet201, and EfficientNetB7) outperforms the state-of-the-art systems, achieving 94.6\% and 96.2\% for the training/testing settings of 10/90 and 20/80, respectively.
For the second criteria, i,e, low-complexity RSIC models ($<$ 5 M trainable parameters) shown in Table~\ref{table:res_04}, our proposed student with knowledge distillation also outperforms the state-of-the-art systems on both training/testing split arrangements, yielding results of 93.3\% for a 10/90 split ratio and 94.8\% for a 20/80 split ratio. 

\section{Conclusion}
This paper has presented, explored, and developed a range of deep convolutional neural networks for the remote sensing image classification (RSIC) task, and in particular considered model complexity.
Through experimentation on the NWPU-RESISC45 benchmark, we obtained two RSIC systems: (1) a teacher developed by combining ConvNeXtLarge, DenseNet201, and EfficientNetB7 network architectures and;  (2) a low complexity student (just 4.7 M trainable parameters), which leverages EfficientNetB0 via knowledge distillation from the teacher. 
Our proposed RSIC systems outperform the state of the art, whether complexity is constrained or not. 
Additionally, a wide range of low- to very low-complexity models using variants of EfficientNetB0 are proposed and explored, which are feasible to apply on edge devices with differing degrees of computational constraint.



\bibliographystyle{IEEEtran}
\bibliography{refs}

\end{document}